# *Research Note*

# On Deducing Conditional Independence from $d$-Separation in Causal Graphs with Feedback

**Radford M. Neal**　　　　　　　　　　　　　　　　　　　　　　　　RADFORD@CS.UTORONTO.CA
*Department of Statistics and Department of Computer Science*
*University of Toronto, Toronto, Ontario, Canada*
`http://www.cs.utoronto.ca/~radford/`

## Abstract

Pearl and Dechter (1996) claimed that the $d$-separation criterion for conditional independence in acyclic causal networks also applies to networks of discrete variables that have feedback cycles, provided that the variables of the system are uniquely determined by the random disturbances. I show by example that this is not true in general. Some condition stronger than uniqueness is needed, such as the existence of a causal dynamics guaranteed to lead to the unique solution.

Causal networks (also known as Bayesian networks or belief networks) are a formalism for representing the joint distribution of a collection of random variables in terms of the conditional distributions for each variable given values for its "parent" variables. The structure of the distribution is represented graphically by a network in which nodes represent variables and arrows are drawn from parent nodes to child nodes. These arrows typically correspond to causal relationships. In the standard formulation, the network is not allowed to have directed cycles.

When a distribution is specified by such a network, the $d$-separation criterion allows one to determine that one set of random variables, $A$, is conditionally independent of another set of random variables, $B$, given values for a third set of random variables $C$. This criterion involves only the presence or absence of arrows in the network, not the detailed numerical specification of the conditional distributions. See Pearl (1988) for a detailed discussion.

Pearl and Dechter (1996) have attempted to extend this framework to networks that may contain directed cycles, which correspond to feedback relationships among variables. When cycles exist, the joint distribution is no longer specified in terms of the product of conditional distributions for children given parents, but rather by saying how the values of the observable variables, $X_1, \ldots, X_n$, are determined by the values for a set of unobserved random disturbances, $U_1, \ldots, U_n$, which are assumed to be independent of each other and to have specified distributions. For each variable, $X_i$, an equation is given specifying that it is equal to some function of the corresponding $U_i$ and of some set of parent variables from among the $X_j$ with $j \neq i$. As before, parent-child relationships are represented graphically by drawing edges with arrows from parent nodes to child nodes.

In order to make this scheme well-defined, Pearl and Dechter require that for any values of $U_1, \ldots, U_n$ there is exactly one set of values for $X_1, \ldots, X_n$ for which all the equations are satisfied. If this uniqueness condition is satisfied, a distribution over $U_1, \ldots, U_n$ will define a distribution over $X_1, \ldots, X_n$. One can then ask what conditional independence properties this distribution might possess.

According to Theorem 2 of Pearl and Dechter (1996), if the $X_i$ are all discrete, the variables $A$ are conditionally independent of the variables $B$ given the variables $C$ if the variables $C$ $d$-separate the variables $A$ and $B$. The $d$-separation criterion can be expressed in terms of the following manipulations of the graph with nodes corresponding to the $X_i$ and with arrows from parents to children:

1) Delete all nodes from the graph except those in $A$, $B$, or $C$ and their ancestors.

2) Connect by an edge every pair of nodes that share a common child.

3) Remove arrows from all the edges — i.e., replace each directed edge by an undirected edge.

If, in the resulting graph, all paths from a node in $A$ to a node in $B$ pass through a node in $C$, then $C$ $d$-separates $A$ from $B$.

Figure 1 shows an example of a distribution defined in this way, which serves as a counterexample to the claim that $d$-separation implies conditional independence for any network satisfying the uniqueness condition. The variables in this example all take values of 0 or 1. The $U_i$ are independent and are equally likely to be 0 or 1. The $X_i$ satisfy the equations shown, in which addition and multiplication are done modulo 2 (i.e., in $Z_2$). Note that $U_2$, $U_3$, $U_6$, and $U_7$ do not appear in the equations, and hence play no role in defining the distribution for $X_1, \ldots, X_7$.[1]

The network and the equations clearly have the required syntactic form. To show that this is a valid example, it is also necessary to show that the $U_i$ uniquely determine values for the $X_i$. One can easily confirm that for any values of the $U_i$ the following values for the $X_i$ will satisfy all the equations:

$$\begin{aligned} X_1 &= U_1 \\ X_2 &= U_4 + U_5 \\ X_3 &= U_4 + U_5 + U_1 \\ X_4 &= U_4 \\ X_5 &= U_5 \\ X_6 &= 0 \\ X_7 &= 0 \end{aligned}$$

To see that this is the only set of values for the $X_i$ that satisfy all the equations, note first that $X_2 + X_4 + X_5$ must be 0, since if it is instead 1, then $X_6 = X_7 + 1$ and $X_7 = X_6$, which is impossible. Hence $X_6 = X_7 = 0$. Since $X_4 = U_4$ and $X_5 = U_5$, we also see that $X_2 + X_4 + X_5 = 0$ implies that $X_2 = U_4 + U_5$, from which it follows that $X_3 = U_4 + U_5 + U_1$, since $X_1 = U_1$.

---

1. The example could have been simplified a bit by omitting $U_1$ and $X_1$ as well, but I have kept them in order to show that the counterexample does not depend on use of such a degenerate network. One can easily make the example less degenerate still by refining the state variables, splitting state 0 into states 0 and $0'$ and state 1 into states 1 and $1'$. The currently unused $U_i$ can then be allowed to influence the choice between 0 and $0'$ and between 1 and $1'$, with this choice having no effect on the other nodes.

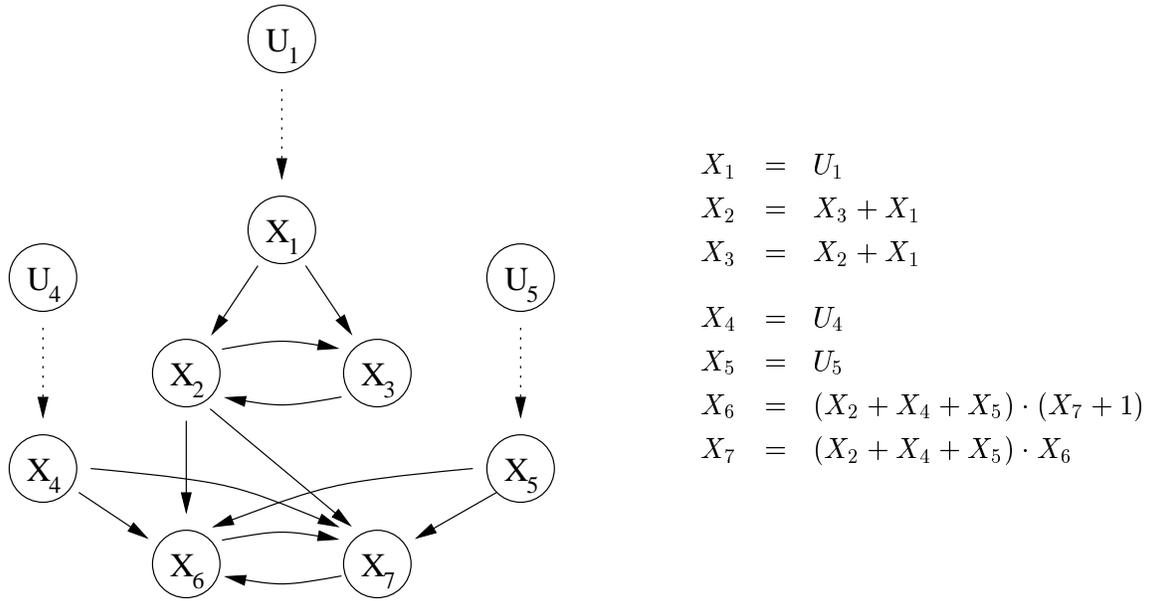

Figure 1: Graphical structure and equations for the counterexample. All variables take values in $\{0, 1\}$. The $U_i$ are independent, with equal probabilities for 0 and 1. Addition and multiplication is done modulo 2. The $U_i$ and the dotted arrows are not formally part of the graph, but are shown for clarity.

Let us now consider whether or not $X_4$ is conditionally independent of $X_5$ given $X_2$. We can see that $X_4$ and $X_5$ are $d$-separated by $X_2$, since in step (1) above, we will delete $X_6$ and $X_7$ and all the edges connecting to them, leaving no path from $X_4$ to $X_5$. However, given any value for $X_2$, the variables $X_4$ and $X_5$ are in fact dependent. As seen above, any set of values for the $X_i$ satisfying the equations must be such that $X_2 + X_4 + X_5 = 0$. Hence, if $X_2 = 0$, then it must be that $X_4 = X_5$, and if instead $X_2 = 1$, then it must be that $X_4 = X_5 + 1$. Conditional on a value for $X_2$, we thus see that $X_4$ is determined by $X_5$, showing that they are not conditionally independent.

The problem appears to arise because even though specifying values for all the $U_i$ uniquely determines values for all the $X_i$, specifying a value for $U_1$ alone leaves *two* sets of values for $(X_1, X_2, X_3)$ that satisfy the equations associated with these variables, even though only one set of values will satisfy the entire set of equations. Which values for $(X_1, X_2, X_3)$ are part of the overall solution depends on the value of $U_4 + U_5$, and this induces a dependence between $X_4 = U_4$ and $X_5 = U_5$ when the value of $X_2$ is known. The removal of part of the graph in step (1) of the procedure for determining $d$-separation eliminates any possibility of accounting for this dependence.

The problem with Pearl and Dechter's proof appears connected with this. They say,

> At this point we invoke the fact that the constraints of C are not arbitrary but are functional, namely, for every values of [ the parents of $x_i$ ] and $u_i$ there is a solution for $x_i$. This implies that, for any set $W$ of variables, the equations associated with non-ancestors of $W$ do not constrain the permitted values of $W$...

The example here shows that the equations involving non-ancestors of $W$ can indeed constrain the permitted values for $W$.

Judea Pearl (personal communication) has suggested that the $d$-separation criterion can be salvaged by requiring not only that $U_1, \ldots, U_n$ uniquely determine $X_1, \ldots, X_n$, but also that this unique solution for $X_1, \ldots, X_n$ can be obtained by a procedure in which the $X_i$ are updated in accordance with the causal structure of the network. In such a casual dynamical procedure, each $X_i$ is repeatedly replaced by the value computed for it from the corresponding $U_i$ and the current values of its parents, according to the equation for that $X_i$, until a stable state is reached. The flow of information in such a procedure follows the direction of the arrows in the network. Consequently, nodes that are not ancestral to any node of interest can have no influence on these nodes, justifying their elimination in step (1) of the procedure for determining $d$-separation.

In general, whether or not such a dynamical procedure eventually finds the solution for $X_1, \ldots, X_n$ may depend on whether the $X_i$ are updated simultaneously or sequentially, and if they are updated sequentially, on the order of these updates. For present purposes, it is sufficient that *some* updating scheme exist that is guaranteed, for any values of $U_1, \ldots, U_n$, to lead to the unique solution, starting with any initial values for $X_1, \ldots, X_n$. If such an update order exists, $d$-separation will imply conditional independence. In the example above, any updating order will for some initial state lead to cyclic behaviour in which the values of $X_6$ and $X_7$ flip back and forth, so the example does not satisfy this stronger condition.

One should note that the example in this note does not invalidate the result of Spirtes (1995) that $d$-separation can be used to determine conditional independence in linear networks of normally distributed variables even if they contain cycles. In the same paper, Spirtes also gave a counterexample showing that $d$-separation need not imply conditional independence in non-linear networks of continuous random variables that contain cycles. This problem cannot be avoided by simply discretizing the continuous variables, as the problem reappears in the form of non-existence or non-uniqueness of solutions. This note shows that in non-linear networks of discrete variables a stronger condition than uniqueness is required for $d$-separation to be valid. Although such a stronger condition involving causal dynamics can be seen as natural, the need to verify this stronger condition does reduce the attractiveness of networks with cycles as a way of formalizing causal situations with feedback.

## Acknowledgments

I thank Peter Spirtes and Thomas Richardson for helpful comments. This work was supported by the Natural Sciences and Engineering Research Council of Canada and by the Institute for Robotics and Intelligent Systems.